\pdfoutput=1

\documentclass[11pt]{article}

\usepackage[final]{acl}
\usepackage{booktabs}
\usepackage{times}
\usepackage{latexsym}
\usepackage{hyperref}
\usepackage{tcolorbox}
\usepackage{mdframed}       
\usepackage{multirow}
\usepackage[T1]{fontenc}

\usepackage[utf8]{inputenc}

\usepackage{microtype}

\usepackage{inconsolata}

\usepackage{graphicx}

\title{Let Me Speak Freely? A Study on the Impact of \\ Format Restrictions on Performance of Large Language Models}

\author{
    Zhi Rui Tam$^1$\thanks{Equal contribution} ,
    Cheng-Kuang Wu$^1$\footnotemark[1],
    Yi-Lin Tsai$^1$,
    Chieh-Yen Lin$^1$,\\
    \textbf{Hung-yi Lee}$^2$,
    \textbf{Yun-Nung Chen}$^{2}$\\
    $^1$Appier AI Research, $^2$National Taiwan University
}

\begin{document}
\maketitle
\begin{abstract}
Structured generation, the process of producing content in standardized formats like JSON and XML, is widely utilized in real-world applications to extract key output information from large language models (LLMs).
This study investigates whether such constraints on generation space impact LLMs’ abilities, including reasoning and domain knowledge comprehension.
Specifically, we evaluate LLMs’ performance when restricted to adhere to structured formats versus generating free-form responses across various common tasks.
Surprisingly, we observe a significant decline in LLMs’ reasoning abilities under format restrictions.
Furthermore, we find that stricter format constraints generally lead to greater performance degradation in reasoning tasks. Our code and results are available online.\footnote{\url{https://github.com/appier-research/structure-gen}}

\end{abstract}

\section{Introduction}

\begin{figure}[t!]
  \includegraphics[width=\columnwidth]{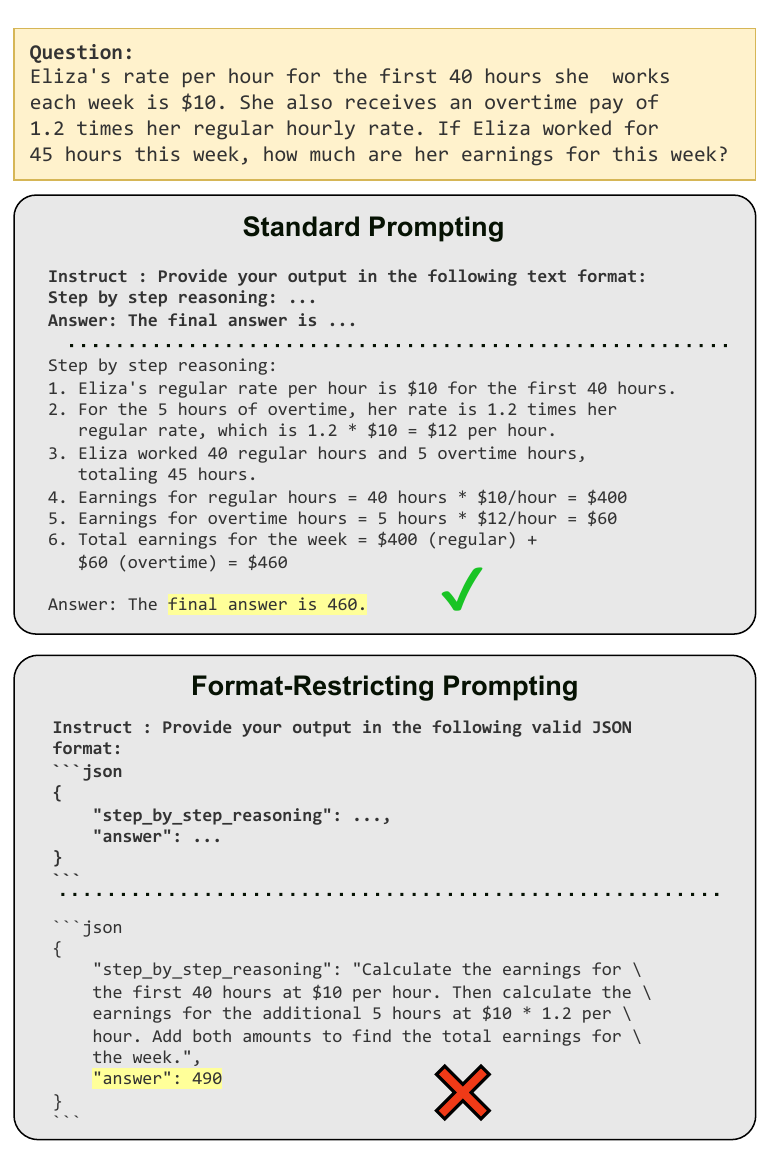}
  \caption{GPT-3.5-turbo prompted with GSM8K math questions in standard natural language answered correctly, but failed when format restrictions were applied.}
  \label{fig:showcase}
\end{figure}

The few-shot in-context learning \cite{brown2020language} and instruction-following \cite{wei2021finetuned} capabilities of large language models (LLMs) have allowed them to solve downstream tasks out of the box.
However, a major obstacle to incorporating LLMs into industrial applications is their lack of adherence to standardized output formats.
This inconsistency complicates output parsing and undermines the reliability of these models.

One common approach to overcoming this obstacle is \textit{structured generation}, which involves providing output in standardized formats like JSON or XML through \textit{format restrictions}.
These restrictions can be implemented in various ways, such as instructing LLMs to adhere to specified formats with \textit{format-restricting instructions}, or using industrial solutions like JSON mode~\cite{oai2024json, gemini2024json}, Instructor~\cite{liu_instructor_2024}, or Guardrails~\cite{prefecthq_marvin_2024}.
These strategies simplify parsing workflows and streamline the integration of LLMs into real-world applications.

Due to the growing demand for structured generation, the research community has shown increased interest in investigating LLMs’ format-following abilities.
For example, IFEval~\cite{zhou2023instruction}, INFOBENCH~\cite{qin2024infobench}, and FOFO~\cite{xia2024fofo} focus on evaluating LLMs’ instruction-following capabilities, including format adherence.
However, these studies do not address a critical question for industrial applications: \textit{Do format-restricting instructions affect the quality of LLMs’ \textbf{generated content}?}
In other words, they fail to explore whether format restrictions degrade LLMs’ performance, which has great business impacts.
This performance degradation is shown in Figure~\ref{fig:showcase}.

In this work, we address the aforementioned research question through extensive empirical experiments.
We present a comprehensive analysis of the potential impacts of format-restricting instructions on LLMs’ performance across a wide range of tasks.
The formats studied include commonly used schemas such as JSON, XML, and YAML.
To the best of our knowledge, this is the first systematic investigation into the relationship between format-restricting instructions and the quality of generated content.
Our contributions are twofold:
\begin{itemize}
    \item We observe declines in LLMs’ reasoning abilities under format restrictions, with stricter constraints generally leading to greater performance degradation in reasoning tasks.
    \item We offer insights into why performance degrades due to format constraints and propose simple approaches to mitigate these issues, thereby achieving both consistent formats and optimal performance.
    \item We explore not only JSON but also other commonly used schemas like XML and YAML. Additionally, we test three different format-restricting strategies: constrained decoding, format-restricting instructions, and NL-to-Format, all of which are applicable to industrial settings.
\end{itemize}


\section{Methodology for Structured Generation} 

To study different levels of format restrictions on downstream performance, we adopt the following three common methodologies in our experiments:

\noindent \textbf{Constrained Decoding (JSON-mode):}
Constrained decoding is a technique that limits the output of LLMs by enforcing predefined token space during the generation process.
Among mainstream LLM providers, \textbf{JSON mode} is a widely implemented instance of this technique, especially due to its extensive use in industrial settings.
This mode, available as a hyperparameter flag in OpenAI and Gemini APIs, ensures the output is valid JSON.
It is assumed that the implementation is similar to the constrained decoding methods described by \citep{willard2023efficient, Koo2024AutomatabasedCF}, and provided in Text-Generation-Inference\footnote{\url{https://github.com/huggingface/text-generation-inference}}.

\noindent \textbf{Format-Restricting Instructions (FRI):}
They direct the LLM to generate responses in standardized formats such as JSON, XML, and YAML, adhering to specified schemas.
These instructions ensure that the generated output follows a structured format, facilitating the extraction and evaluation of the final answer.
This approach is more relaxed than constrained decoding, as it does not enforce a predefined token space.

\noindent \textbf{NL-to-Format:}
This two-step process first instructs the LLM to answer the question in natural language, and then instructs it to convert its response into the target format schema.
As the most relaxed version of structured generation, this method decouples \textit{content generation} from \textit{format adherence}, aiming to maintain the performance of unrestricted natural language responses while still providing structured output.

\section{Experiments}
\begin{figure}[t!]
  \includegraphics[width=\columnwidth]{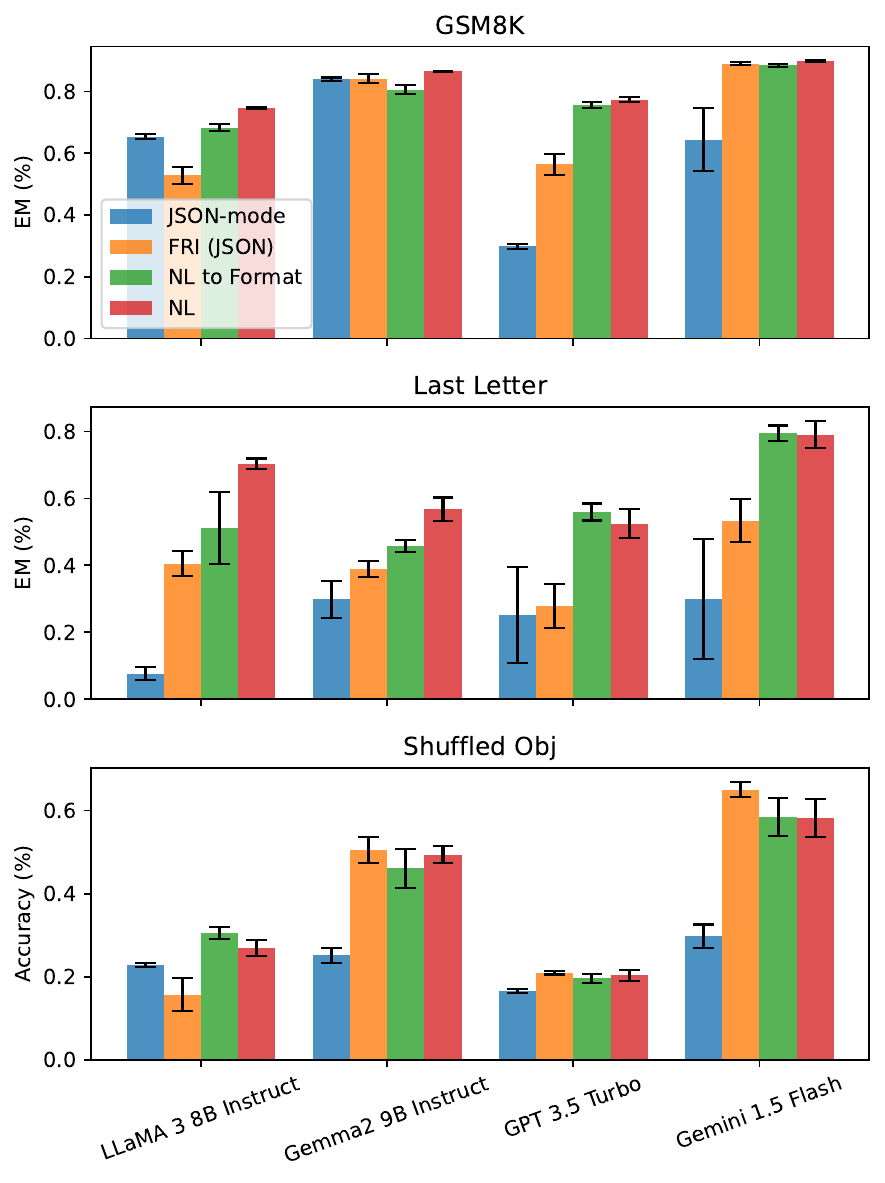}
  \caption{When comparing reasoning related task such as GSM8K, Last Letter and Shuffled Objects, we found more relaxed prompts typically yields better results as JSON-mode performs the worse in most case followed by FRI, NL to Format and Natural Language (NL)}
  \label{fig:reasoning_restriction}
\end{figure}

\begin{figure*}[ht]  \includegraphics[width=\linewidth]{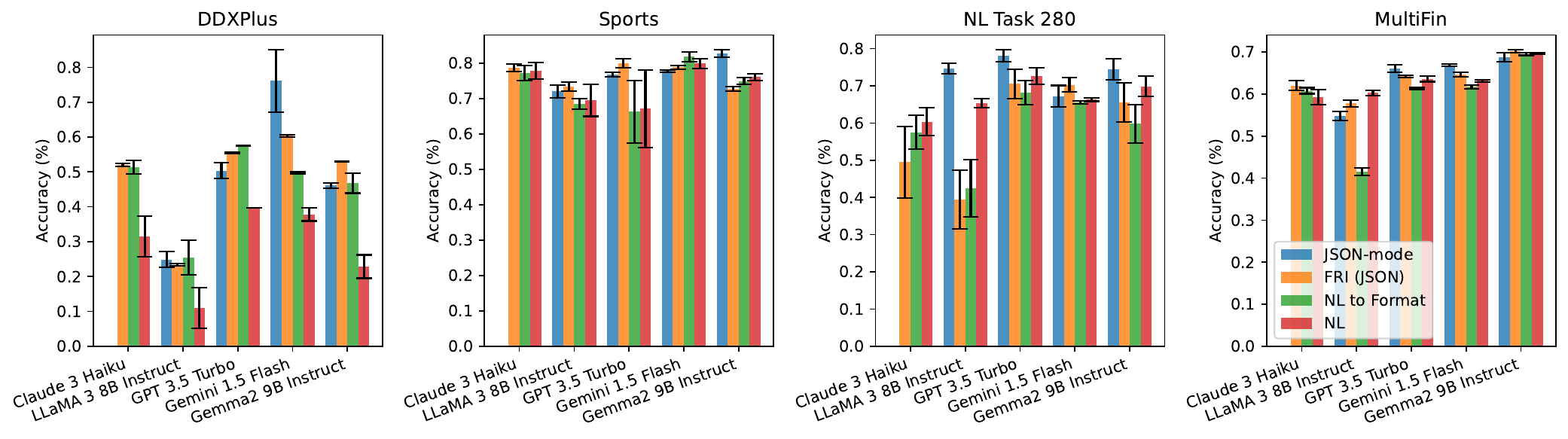}
  \caption{Classification related tasks on DDXPlus, Sports, Task280 and Multifin in different levels of format restriction.}
  \label{fig:classification_restriction}
\end{figure*}

\subsection{Datasets}
We adopt datasets from various domains, categorized by the primary skills they assess:

\subsubsection{Reasoning Tasks}
\noindent \textbf{GSM8K} \citep{cobbe2021gsm8k}: A collection of mathematical problems set in natural language contexts, reflecting daily life scenarios. This dataset challenges LLMs to generate necessary intermediate reasoning steps.

\noindent \textbf{Last Letter Concatenation} \citep{wei2022chain}: This task requires LLMs to produce a string by concatenating the last letters of a sequence of words, testing their ability to perform symbolic reasoning.

\noindent \textbf{Shuffled Objects} \citep{ghazal2013bigbench}: This evaluate set from BigBench evaluates the ability to infer the final state given an initial state and a sequence of shuffling events. We use the entire validation set in our experiments.

\subsubsection{Classification Tasks}

\noindent \textbf{DDXPlus} \citep{tchango2022ddxplus}: A multiple-choice medical diagnosis dataset where LLMs must select the most appropriate diagnosis from 49 possible diseases based on a given patient profile. We use a subset provided by StreamBench \citep{Wu2024StreamBenchTB} due to the extensive number of questions.

\noindent \textbf{MultiFin} \citep{Jrgensen2023MultiFinAD}: A multi-choice financial dataset that requires classifying a given paragraph into one of five categories.

\noindent \textbf{Sports Understanding} \citep{ghazal2013bigbench}: This task from BigBench tests LLMs' ability to determine whether an artificially constructed sentence relating to sports is plausible or implausible.

\noindent \textbf{NI - Task 280} \citep{naturalinstructions}: A multiple-choice stereotype classification task based on a given paragraph. We included this task as it has been found to be sensitive to change in prompt formatting, with performance variations of up to 56\% \citep{sclar2023quantifying}.

\subsection{Output Format}

When designing the output format for each format, we wish to keep the schema simple; hence, we limit the number of key-value pairs for each dataset to 2: reasoning and answer fields. On top of this limitation, we permute the naming of the field names (e.g., "reasoning", "step-by-step reasoning").

While the outputs in our study may appear simplistic, converting Large Language Model (LLM) responses to a desired format is not trivial in practice. LLMs' output often deviates from instructions, necessitating complex parsing code to handle various response variations and edge cases, particularly when separating reasoning from the final answer. This problem is exacerbated when switching between different LLMs, as each model may have its own preferred output format, potentially breaking existing parser code. We have encountered this issue numerous times when building LLM applications, often resorting to instructing LLMs to respond in structured formats (e.g., JSON) to reduce the complexity of our parser code.

Our choice of simple output structures (one reasoning and one final answer field) was deliberate, allowing us to focus on the impact of structural bias on LLM reasoning ability, which is the primary aim of our work. We acknowledge that exploring LLM robustness with more complex output structures would be valuable. We have noted this as an important direction for future research.

\subsection{Model}
For all experiments, we compare \textit{gpt-3.5-turbo-0125} ~\cite{Achiam2023GPT4TR}, \textit{claude-3-haiku-20240307} \citep{claude2024anthropic}, \textit{gemini-1.5-flash} \citep{team2023gemini}. For open weights model we use \textit{LLaMA-3-8B-Instruct} \citep{llama2024meta} and \textit{Gemma-2-9B-Instruct} \citep{team2024gemma} inference using Text-Generation-Server for its support in \textbf{JSON mode}\footnote{\url{https://github.com/huggingface/text-generation-inference/pull/1938}}.

\subsection{Evaluation method}

\noindent \textbf{Metrics.}
To assess the performance of the models across the diverse range of tasks, we employ task-specific evaluation metrics. For the classification-based tasks (Sports Understanding, DDXPlus, Natural Instruction Task 280, and MultiFin), we use accuracy as the primary metric. For the Last Letter Concatenation and GSM8K, we utilize the exact match metric where the final answer must be the extact string match with the actual answer.

\noindent \textbf{Perfect Text Parser.}
To disentangle format errors from the actual performance of the generated content, we use an LLM prompted to extract the final answer from the text, rather than relying on regex or string parsers.
This approach acts as a perfect parser, minimizing errors introduced when switching between different models.
Our ablation study, comparing different models, found that \textit{claude-3-haiku-20240307} is the most consistent when using \textit{gpt-4-turbo} as a human reference, compared to four other low-cost APIs. Detailed comparison between \textit{gpt-4-turbo} between human parsed answers as well as comparison of other LLMs can be found in Appendix~\ref{app:llm_parser_ablation}.

\noindent \textbf{Consideration for Prompt Sensitivity.}
Previous studies \citep{chen2023exploring, sclar2023quantifying, zhu2023promptbench, mizrahi2024state} have shown that LLMs are sensitive to slight variations in prompts.
To account for this, we evaluate our approach by nine prompt combinations: three task descriptions and three JSON, XML, and YAML schemas with slight variations in wording or format.
For natural language prompting, we include three variations in text formats (e.g., \textit{Give your reason first followed by your answers}).
Details of the task description prompts and FRI prompts can be found in Appendix \ref{app:prompt_format_details}.

\section{Main Results}

\subsection{Impact of Format Restriction on Final Results}
We investigate the effects of format restrictions on LLM performance by examining three progressively relaxed prompting approaches: JSON-mode, FRI, and NL-to-Format conversion.

We evaluate these approaches on datasets with exact match scores: GSM8K and Last Letter Concatenation presented in Figure \ref{fig:reasoning_restriction}. Surprisingly, JSON-mode performs significantly worse than FRI (JSON) on the Last Letter task. Upon inspection, we found that 100\% of GPT 3.5 Turbo JSON-mode responses placed the "answer" key before the "reason" key, resulting in zero-shot direct answering instead of zero-shot chain-of-thought reasoning.

Comparing NL-to-Format with unrestricted Natural Language responses, we observe nearly identical performance across most models, as both derive answers from the same initial natural language response. However, NL-to-Format occasionally introduces generation errors, leading to slightly lower performance for LLaMA 3 8B Instruct, while other models maintain consistent scores across both settings.

These findings suggest that the degree and implementation of format restrictions can significantly impact LLM performance, particularly in reasoning tasks. The order of keys in structured outputs and the decoupling of reasoning from format adherence emerge as important factors in maintaining LLM capabilities while providing structured responses.

When evaluating classification datasets, we observe a different trend compared to reasoning tasks, as illustrated in Figure \ref{fig:classification_restriction}. Notably, in the DDXPlus dataset, Gemini 1.5 Flash demonstrates a significant performance boost when JSON-mode is enabled. Across other classification datasets, JSON-mode performs competitively, and in some cases, surpasses the other three methodologies.

We hypothesize that JSON-mode improves classification task performance by constraining possible answers resulted in reducing errors in answer selection. Conversely, natural language responses may introduce distractions, leading to parsing errors. These findings suggest format restrictions' impact on LLM performance is task-dependent: stringent formats may hinder reasoning-intensive tasks but enhance accuracy in classification tasks requiring structured outputs.
\section{Discussion}

\subsection{Impact on looser format restriction}

To further investigate the effects of format restrictions, we examine a variation of the Soft Restrict setting where we remove the schema restriction from the prompt description. Instead of providing a specific schema (e.g., \textit{"Reply your answer in JSON format with the following schema: \{ "reason": ..., "answer": ... \}"}), we simply instruct the LLM to output in the target format language (e.g., \textit{"Reply your answer in JSON format."}).
Table \ref{tab:loose_to_strict} illustrates the effects of removing the schema restriction on the GSM8K dataset. We observe significant improvements in average scores and lower standard deviations across different prompt perturbations for Claude 3 Haiku, GPT-3.5 Turbo, and LLaMA 3 8B Instruct. These results suggest that while structured outputs can be beneficial for downstream processing, overly restrictive schemas may hinder LLM performance, particularly in reasoning-intensive tasks. 

This finding suggests that a balance must be struck between the desire for easily parseable, structured outputs and the need to preserve the LLM's inherent reasoning abilities. Practitioners may want to consider using looser format restrictions when dealing with complex reasoning tasks, while still maintaining some level of structure to facilitate downstream processing.

\begin{table}[t]
\centering
\small
\begin{tabular}{@{}lcccc@{}}
\toprule
\textbf{Model} & \textbf{Text} & \textbf{JSON} & \textbf{XML} & \textbf{YAML} \\
\midrule
\underline{\textit{gemini-1.5-flash}}   & 89.33 & \textbf{89.66} & \textbf{89.26} & \textbf{89.21} \\
             & (0.8) & (0.3) & (0.3) & (0.4) \\
+ schema constraint & - & 89.21 & 88.20 & 87.42 \\
          & - & (1.5) & (2.2) & (3.7) \\
\midrule
\underline{\textit{claude-3-haiku}} & 86.51 & \textbf{86.99} & \textbf{86.96} & \textbf{82.89}\\
 & (0.8) & (0.2) & (0.6) & (5.7) \\
+ schema constraint & - & 23.44 &  79.76 &80.63\\
 & - & (22.9) & (7.0) & (2.8)\\
\midrule
\underline{\textit{gpt-3.5-turbo}} & 75.99 & \textbf{74.70} & \textbf{60.45} & 71.58 \\
 & (3.1) & (1.1) & (7.2) & (3.0) \\
+ schema constraint & - & 49.25 & 45.06 &  \textbf{73.85}\\
 & - & (12.0) & (19.9) & (5.6) \\
\midrule
\underline{\textit{LLaMA-3-8B}} & 75.13 & \textbf{64.67} & \textbf{65.07} & \textbf{69.41} \\
 & (0.9) & (2.23) & (0.56) & (0.95) \\
+ schema constraint & - & 48.90 & 56.74 & 46.08 \\
 & - & (6.7) & (8.3) & (16.8) \\
\bottomrule
\end{tabular}
\caption{Comparing results without and with schema constraint, adding schema not only increase the sensitivity to prompt but also degrade in average performance.}
\label{tab:loose_to_strict}
\end{table}
\subsection{Comparison Across Different Formats}
\begin{figure*}[ht]
  \includegraphics[width=\linewidth]{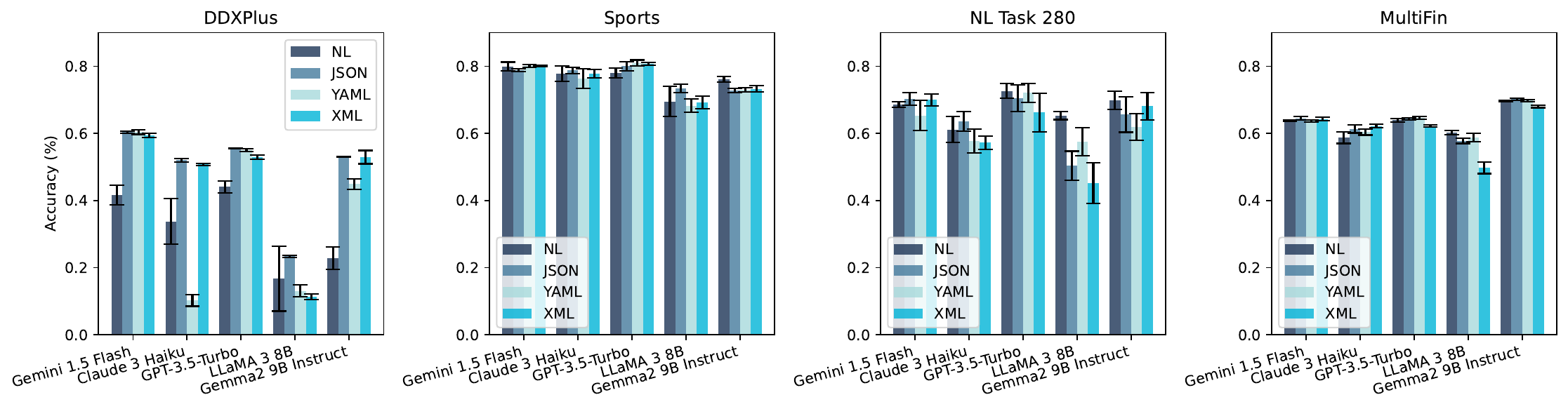}
  \caption{Comparison of different formats in classification related tasks on DDXPlus, Sports, Task280 and Multifin. NL=Natural Language. We showed the averaged accuracy for each format over 9 different prompts with standard deviation error.}
  \label{fig:classification}
\end{figure*}
In this section we ablate the format language by comparing not just JSON but also XML and YAML format. Since all 3 language comes in different grammar syntax rules and restriction. We expect each models might perform differently for example Claude-3-Haiku uses XML for tool use schema.

On hindsight we do not see any structure format which consistency stands out from others which generalized across all models in Figure \ref{fig:classification}. For Gemini model, we found JSON is more consistent however it does not always outperform other format for example Claude-3-Haiku.

In Table \ref{tab:json_mode} we found in classification task JSON-mode performs much better than text due to the restriction on answer space. However in reasoning related task, JSON-mode failed to adhere to the order of reasoning first followed by answer causing a large drop in final performance.

\subsection{Structure Format and Parsing Error Rates}

We initially hypothesized that the performance gap between text and structured formats might be attributed to parsing errors during answer extraction. However, our analysis of error rates across different formats and models, as shown in Table \ref{tab:parse_err}, reveals that this is not the primary factor. In fact, Gemini 1.5 Flash and GPT 3.5 Turbo exhibit near zero parsing failures in all three formats. In the LLaMA 3 8B setting, the parsing error rate for the Last Letter task in JSON format is only 0.148\%, yet there exists a substantial 38.15\% performance gap as seen in Table \ref{tab:loose_to_strict}.

This finding suggests that the performance differences between formats are not primarily due to parsing errors, but rather to the impact of format restrictions on the LLM's reasoning and generation processes. However, we discovered that parsing errors, when present, can be effectively mitigated through a simple corrective step.

\begin{figure}[t!]
  \includegraphics[width=\columnwidth]{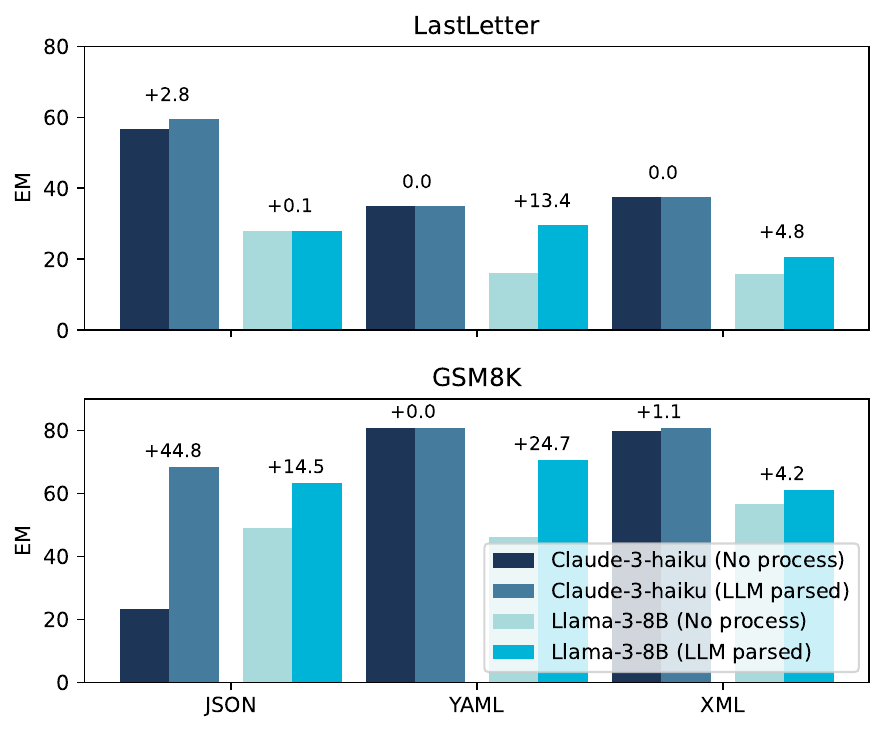}
  \caption{We found high parsing errors in Table \ref{tab:parse_err} can be patched by calling a second prompt to fix any syntax error found in the previous response.}
  \label{fig:error_fixed}
\end{figure}

By prompting Claude-3-Haiku to reformat any output with parsing errors for both Claude 3 Haiku and LLaMA 3 8B (the two models with the highest percentage of parsing errors), we observed improved scores in JSON and YAML formats, as illustrated in Figure \ref{fig:error_fixed}. This approach demonstrates the potential for enhancing the reliability of structured outputs without sacrificing the benefits of format-specific optimizations.

\subsection{Study on Structure Generation with Context-free Grammars}

A newer revision of the model \textit{gpt-4o-mini-2024-07-18} now supports Context-free Grammars via a so-called Structure Output API. This API allows users to provide a predefined JSON schema, ensuring the response adheres to it with 100\% guarantee. It's important to note that this differs from the previously mentioned JSON-mode on OpenAI's models, which uses the OpenAI function calling API. We conducted experiments on 3 reasoning datasets using gpt-4o-mini, denoting the newer structured output method as JSON-schema. Results are shown in Table \ref{tab:gpt_4o_mini_perf}.

\begin{table}[t]
\centering
\small
\setlength{\tabcolsep}{4pt}
\begin{tabular}{@{}lcccc@{}}
\toprule
\textbf{Task} & \textbf{NL} & \textbf{FRI} & \textbf{JSON-Mode} & \textbf{JSON-Schema} \\
\midrule
\underline{\textit{GSM8K}} & \textbf{94.57} & 87.17 & 86.95 & 91.71 \\
 & (3.95) & (4.43) & (1.36) & (0.68) \\
\midrule
\underline{\textit{Shuffle Obj}} & \textbf{82.85} & 81.46 & 76.43 & 81.77 \\
 & (5.67) & (3.71) & (9.74) & (6.86) \\
\midrule
\underline{\textit{Last Letter}} & 83.11 & 84.73 & 76.00 & \textbf{86.07} \\
 & (3.54) & (2.99) & (6.69) & (3.33) \\
\bottomrule
\end{tabular}
\caption{Performance of \textit{gpt-4o-mini-2024-07-18} across tasks and formats. In 2 out of 3 reasoning datasets, NL (Natural Language) still performs slightly better than JSON-Schema.}
\label{tab:gpt_4o_mini_perf}
\end{table}

\begin{figure}[t!]
  \includegraphics[width=\columnwidth]{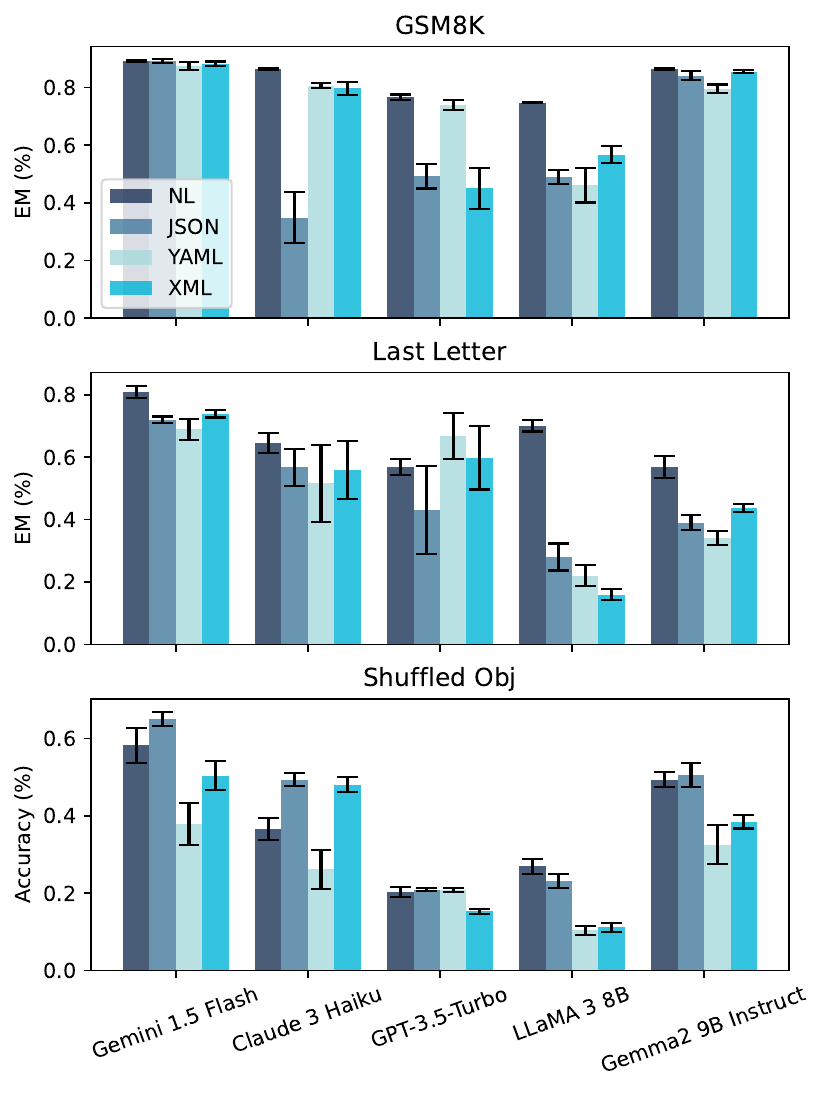}
  \caption{Comparison of JSON, YAML, XML with Natural Language (NL) response on reasoning related task. NL still performs better than other formats with the exception of GPT-3.5-Turbo.}
  \label{fig:reasoning}
\end{figure}

\begin{table*}[t!]
\centering
\caption{Parsing error percentage across different models. We want to highlight that despite having near zero parsing error in Gemini-Flash XML and YAML, there's still degradation in the final benchmark scores.}
\label{tab:parse_err}
\centering
\small
\begin{tabular}{lcccccccc}
\toprule
& \bf Task & \multicolumn{2}{c}{\bf Reasoning} & \multicolumn{4}{c}{\bf Classification} \\
\cmidrule(lr){3-4}
\cmidrule(lr){5-9}
Model & \textbf{Format}  & Last Letter  & GSM8K  & DDXPlus   & Sports & Task280 & MultiFin \\
\midrule
Gemini-Flash & JSON & 0.0 & 0.03 &  0.37 & 0.0 & 0.0 & 0.0  &\\
& XML & 0.0 & 0.19 &  1.26 & 0.0 & 0.22 & 0.0  & \\
& YAML & 0.0 & 0.0 & 0.68 & 0.06 & 6.46 & 0.0  &\\
\midrule
Claude-3-Haiku & JSON & 3.48  & 60.07 & 0.09 & 0.0 & 10.26 & 0.0\\
& XML & 0.0  & 1.85 & 0.48 &  0.0& 0.41 & 0.0 \\
& YAML &0.0  & 0.0 & 86.66 & 1.02 & 0.13 & 0.0 \\
\midrule
GPT-3.5-Turbo & JSON & 0.0  & 0.13 &  0.0& 0.0 & 0.0 & 0.0\\
& XML & 0.0  &  0.24 & 0.35 & 0.0 & 0.0 & 0.0\\
& YAML &0.0  & 0.0 & 0.32 &  1.23& 0.08 & 0.0\\
\midrule
LLaMA 3 8B & JSON & 0.15  &  22.75 &1.63  & 0.28 & 1.61 & 0.0 \\
& XML & 17.93 & 7.62 & 32.45 & 6.54 & 22.04 & 5.78 \\
& YAML & 32.40 &33.18  & 34.40 & 7.16 & 2.19 & 0.14 \\
\bottomrule
\end{tabular}
\end{table*}

\section{Related Work}
Our study can be summarized into two genres : reasoning ability of LLM and format following.

In study of LLMs reasoning ability, early work by \cite{kojima2022large} found using "Think step-by-step" can  elicit reasoning ability without few shot examples. Subsequent study \citep{jin2024impact} shows that the number of reasoning steps correlates with the final accuracy. Recent work by \citep{Wang2024ChainofThoughtRW} found Chain-of-Thought (CoT) reasoning seed prompt \cite{kojima2022large} can be removed with a carefully crafted CoT decoding schema. 

The exploration of LLMs' ability to follow instructions and produce responses in specified formats was first addressed by IFEval~\cite{zhou2023instruction} which aimed to evaluate the general instruction-following ability of LLMs, and it contains a subset of test instances specifically assessing format-following. INFOBENCH~\cite{qin2024infobench} introduces a broader coverage of instructions and conducts a more fine-grained analysis by decomposing the instructions into different categories, including format specifications. FOFO~\cite{xia2024fofo} is a benchmark solely focused on the format-following ability of LLMs. However, these works do not explore if format instruction interfere with downstream performance.

\section{Conclusion}

Our study reveals that structured generation constraints significantly impact LLM performance across various tasks. Format restrictions, particularly constrained decoding (JSON-mode), can hinder reasoning abilities while enhancing classification task accuracy. Looser format restrictions generally improve performance and reduce variance in reasoning tasks. Parsing errors, while not the primary cause of performance differences, can be mitigated through corrective prompting. These findings underscore the importance of balancing format adherence, reasoning capabilities, and cost efficiency in LLM applications. Given that our study focuses on reasoning-intensive tasks, future work should explore how reasoning tasks of varying difficulty, from intensive to simple, are affected by restrictive formats and LLMs. To mitigate the performance degradation of LLMs due to restrictive formats, future studies should include a wider range of training data that contains instructions in various restrictive formats in local LLMs.

\bibliography{references}

\appendix

\section{Limitation}

This study contains two primary limitations. First, due to cost constraints, we were unable to include results from more powerful language models such as LLaMA 70B or GPT-4o in our experiments. The inclusion of these models could potentially provide additional insights into how performance scales with model size and architecture. Second, our evaluation dataset, while diverse, is limited in scope. A broader range of tasks and domains could offer a more comprehensive assessment of the proposed approach's effectiveness and generalizability.

\label{sec:appendix}
\section{Choosing which LLMs as answer extraction}
\label{app:llm_parser_ablation}

We first validate if existing LLMs such as \textit{gpt-4-turbo} can the perfect parser in answer extraction in reasoning tasks such as GSM8K, Last Letter Concatenation. We sampled 300 responses in total: 100 each from Last Letter, Shuffle Object, and GSM8K, each of the responses were independently parsed by human evaluators. We then compared the human-parsed answers with those extracted by GPT-4-turbo. The result shown in Table \ref{tab:parser-evaluation}, shows \textit{gpt-4-turbo} can indeed denote as a perfect parser in these 3 cases.

\begin{table}[t]
\centering
\small
\begin{tabular}{lcr}
\toprule
\textbf{Task} & \textbf{Examples} & \textbf{Accuracy (\%)} \\
\midrule
Last Letter & 100 & 97.0 \\
Shuffle Obj & 100 & 96.0 \\
GSM8K & 100 & 100.0 \\
\midrule
\textbf{Average} & 300 & 97.7 \\
\bottomrule
\end{tabular}
\caption{Alignment between GPT-4-Turbo and human annotations across different tasks.}
\label{tab:parser-evaluation}
\end{table}

\begin{figure}[t!]
  \includegraphics[width=\columnwidth]{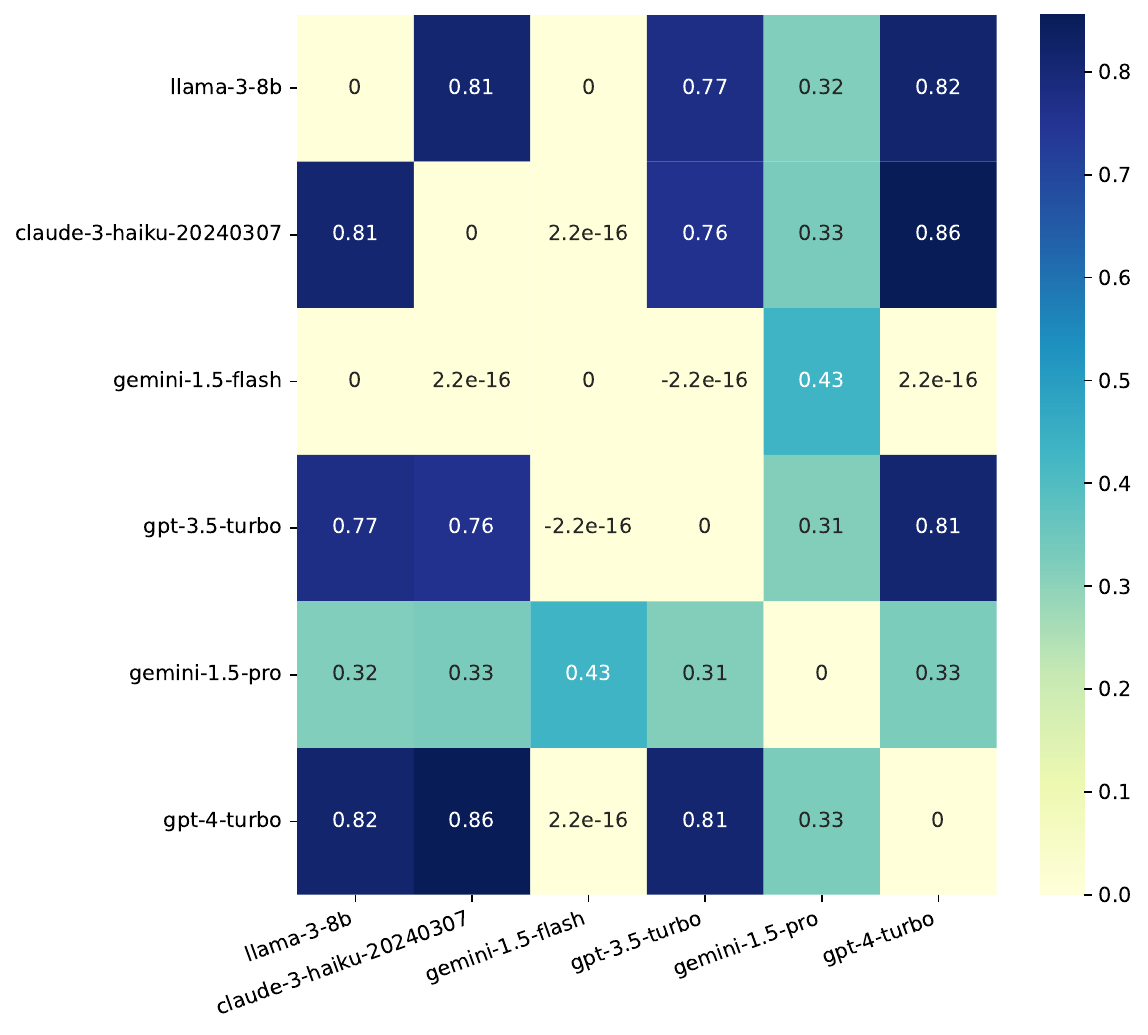}
  \caption{Agreement scores among all LLMs on the final extracted answes.}
  \label{fig:llm_parser_ablation}
\end{figure}

To select the best and low cost answer LLM parser, we select 200 samples from six datasets response in natural language format which a total of 1,200 samples. We then use \textit{gpt-4-turbo} as best LLM answer parser as our reference and calculate the kappa cohen score with 3 LLMs candidates: \textit{gemini-1.5-flash}, \textit{claude-3-haiku-20240307} and \textit{llama-3-8b-instruct} in Figure \ref{fig:llm_parser_ablation}. Result shows \textit{claude-3-haiku-20240307} has the highest aggreement with \textit{gpt-4-turbo} at 0.86 followed by \textit{llama-3-8b-instruct}.

\section{Cost Comparison Across Different Formats}

An important consideration in deploying LLM applications in industry settings is the associated token cost. We analyzed the input and output tokens across our experiments for all models and formats. For brevity, we present the averaged results from all six datasets in Table \ref{tab:avg-format-comparison}.
Our analysis reveals that text and YAML formats generally incur similar costs. Interestingly, we found that YAML is the most cost-effective format for LLaMA-3-8B, Gemini-1.5-Flash, and GPT-3.5-Turbo. Surprisingly, for Claude-3-Haiku, the lowest cost is associated with the text format, which is unexpected given the prevalence of XML examples in their documentation for tool use. The full cost breakdown for each dataset can be found in Table \ref{tab:model_cost_breakdown}, providing a more detailed view for practitioners interested in fine-tuning their approach for specific use cases.

\begin{table}[t]
\centering
\small
\begin{tabular}{lcccc}
\toprule
\textbf{Model} & \textbf{text} & \textbf{json} & \textbf{xml} & \textbf{yaml} \\
\midrule
LLaMA-3-8b & 0.11 & 0.09 & 0.09 & 0.08 \\
Gemini-1.5-Flash & 0.20 & 0.21 & 0.21 & 0.19 \\
Claude-3-Haiku & 0.20 & 0.30 & 0.30 & 0.29 \\
GPT-3.5-Turbo & 0.35 & 0.23 & 0.24 & 0.23 \\
\bottomrule
\end{tabular}
\caption{Comparison of total costs (US dollar per 1000 entries) for different models and output formats. Numbers are averaged over all 6 datasets. }
\label{tab:avg-format-comparison}
\end{table}

\begin{table*}[t]
\centering
\small
\begin{tabular}{llcccccccccccc}
\toprule
& & \multicolumn{3}{c}{gemini-1.5-flash} & \multicolumn{3}{c}{llama-3-8b} & \multicolumn{3}{c}{claude-3-haiku} & \multicolumn{3}{c}{gpt-3.5-turbo} \\
\cmidrule(lr){3-5} \cmidrule(lr){6-8} \cmidrule(lr){9-11} \cmidrule(lr){12-14}
Dataset & Format & In & Out & Tot & In & Out & Tot & In & Out & Tot & In & Out & Tot \\
\midrule
\multirow{4}{*}{lastletter} 
& text   & 0.04 & 0.09 & 0.12 & 0.02 & 0.02 & 0.04 & 0.03 & 0.12 & 0.15 & 0.05 & 0.07 & 0.12 \\
& json   & 0.04 & 0.10 & 0.14 & 0.02 & 0.03 & 0.05 & 0.03 & 0.17 & 0.21 & 0.06 & 0.05 & 0.11 \\
& xml    & 0.04 & 0.10 & 0.14 & 0.02 & 0.03 & 0.05 & 0.03 & 0.15 & 0.18 & 0.06 & 0.07 & 0.13 \\
& yaml   & 0.04 & 0.09 & 0.13 & 0.02 & 0.02 & 0.05 & 0.03 & 0.14 & 0.18 & 0.06 & 0.09 & 0.14 \\
\midrule
\multirow{4}{*}{gsm8k} 
& text   & 0.05 & 0.13 & 0.18 & 0.03 & 0.03 & 0.06 & 0.04 & 0.23 & 0.27 & 0.07 & 0.16 & 0.23 \\
& json   & 0.05 & 0.14 & 0.20 & 0.03 & 0.03 & 0.07 & 0.04 & 0.29 & 0.33 & 0.08 & 0.12 & 0.19 \\
& xml    & 0.06 & 0.14 & 0.19 & 0.03 & 0.03 & 0.07 & 0.05 & 0.27 & 0.32 & 0.08 & 0.12 & 0.20 \\
& yaml   & 0.05 & 0.13 & 0.18 & 0.03 & 0.03 & 0.06 & 0.04 & 0.28 & 0.32 & 0.08 & 0.14 & 0.22 \\
\midrule
\multirow{4}{*}{multifin} 
& text   & 0.05 & 0.01 & 0.06 & 0.03 & 0.00 & 0.03 & 0.03 & 0.02 & 0.05 & 0.07 & 0.02 & 0.09 \\
& json   & 0.05 & 0.02 & 0.07 & 0.03 & 0.00 & 0.03 & 0.04 & 0.05 & 0.09 & 0.07 & 0.02 & 0.09 \\
& xml    & 0.05 & 0.02 & 0.07 & 0.03 & 0.01 & 0.04 & 0.04 & 0.04 & 0.08 & 0.08 & 0.03 & 0.10 \\
& yaml   & 0.05 & 0.01 & 0.06 & 0.03 & 0.00 & 0.03 & 0.04 & 0.02 & 0.06 & 0.07 & 0.01 & 0.08 \\
\midrule
\multirow{4}{*}{sports} 
& text   & 0.04 & 0.04 & 0.08 & 0.02 & 0.01 & 0.03 & 0.03 & 0.10 & 0.13 & 0.05 & 0.05 & 0.10 \\
& json   & 0.04 & 0.06 & 0.10 & 0.02 & 0.01 & 0.04 & 0.03 & 0.11 & 0.15 & 0.06 & 0.07 & 0.12 \\
& xml    & 0.04 & 0.07 & 0.11 & 0.02 & 0.02 & 0.04 & 0.03 & 0.14 & 0.17 & 0.06 & 0.08 & 0.14 \\
& yaml   & 0.04 & 0.05 & 0.08 & 0.02 & 0.01 & 0.04 & 0.03 & 0.12 & 0.15 & 0.05 & 0.06 & 0.11 \\
\midrule
\multirow{4}{*}{task280} 
& text   & 0.04 & 0.05 & 0.09 & 0.03 & 0.01 & 0.03 & 0.03 & 0.05 & 0.08 & 0.06 & 0.04 & 0.11 \\
& json   & 0.05 & 0.04 & 0.08 & 0.03 & 0.01 & 0.03 & 0.04 & 0.07 & 0.11 & 0.07 & 0.04 & 0.11 \\
& xml    & 0.05 & 0.04 & 0.09 & 0.03 & 0.01 & 0.04 & 0.04 & 0.08 & 0.11 & 0.07 & 0.05 & 0.12 \\
& yaml   & 0.04 & 0.03 & 0.07 & 0.03 & 0.01 & 0.03 & 0.04 & 0.05 & 0.09 & 0.06 & 0.03 & 0.10 \\
\midrule
\multirow{4}{*}{ddxplus} 
& text   & 0.26 & 0.15 & 0.41 & 0.15 & 0.04 & 0.18 & 0.19 & 0.20 & 0.38 & 0.38 & 0.21 & 0.59 \\
& json   & 0.22 & 0.18 & 0.41 & 0.13 & 0.06 & 0.19 & 0.19 & 0.33 & 0.52 & 0.34 & 0.15 & 0.48 \\
& xml    & 0.23 & 0.19 & 0.42 & 0.14 & 0.06 & 0.19 & 0.19 & 0.37 & 0.56 & 0.34 & 0.18 & 0.51 \\
& yaml   & 0.22 & 0.15 & 0.37 & 0.13 & 0.05 & 0.18 & 0.19 & 0.31 & 0.50 & 0.33 & 0.15 & 0.48 \\
\bottomrule
\end{tabular}
\caption{Performance comparison of different models across various datasets and formats. Values represent processing times in seconds for Input (In), Output (Out), and Total (Tot).}
\label{tab:model_cost_breakdown}
\end{table*}

\section{Additional models}

\begin{figure*}[ht]  \includegraphics[width=\linewidth]{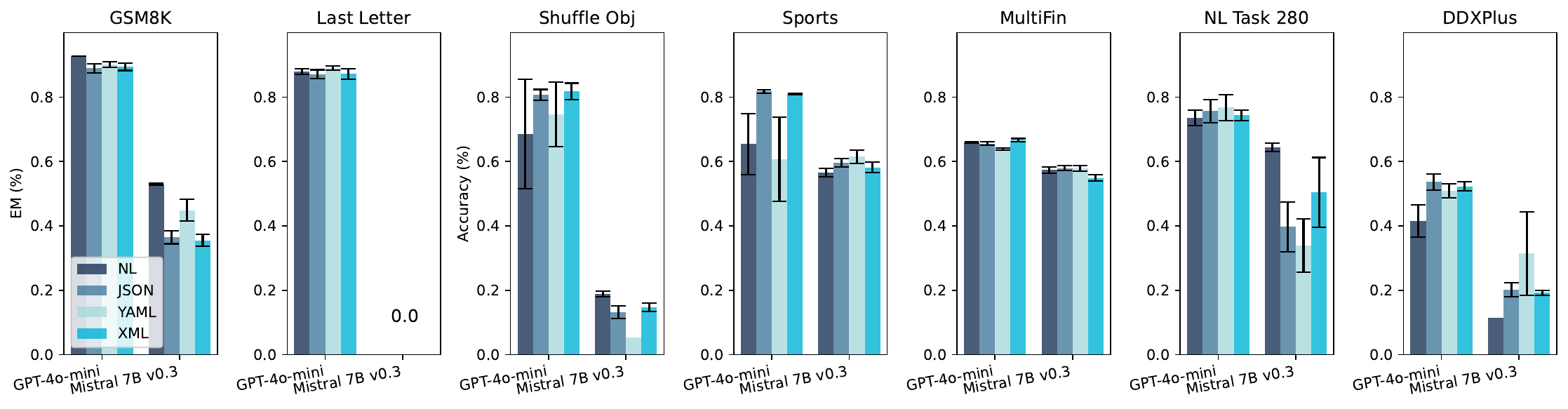}
  \caption{Exact Match scores on GSM8K and Last Letter on reasoning related datasets. Classification related tasks on Shuffled Object, Sports Understanding, MultiFin, NL Task 280 and DDXPlus in different levels of format restriction. In general, we found GPT-4o is quite consistent on adding format restriction. In the Last Letter task, the exact match scores of Mistral-7B-v0.3 across all 4 formats are very close to 0\%, which are not explicitly shown in the figure.}
  \label{fig:extra_results}
\end{figure*}

We also tested additional models from Mistral and OpenAI : \textit{Mistral-7b-v0.3}, \textit{GPT-4o-mini-2024} on format prompt variation in GSM8K, Last Letter, Shuffled Object, Sports Understanding, MultiFin, NL Task 280 and DDXPlus. The result is visualized in Figure \ref{fig:extra_results}.

\section{Comparison between using regex and LLM as answer parser in GSM8K}

To answer the difference between using regex parser to extract the final strict match answer, we calculate the Exact Match score in GSM8K results using the prompt format template "The final answer is". Table \ref{tab:parser-comparison} results reveal a significant gap between regex match and LLM as final answer parser in EM score across various language models, highlighting the limitations of using only one strict regex matching for different models. For example, GPT-3.5-Turbo shows a 31.8 percentage point improvement from regex match (43.7\%) to overall accuracy (75.5\%), while Gemini-1.5-Flash exhibits an even larger 43.5 point difference. This pattern is consistent across all models, with mistral-7b demonstrating the most dramatic 42 point increase.

These disparities underscore the value of using LLMs as answer parsers, as they can understand and evaluate responses beyond literal string matching, accounting for paraphrases and contextual understanding, thus providing a more nuanced and accurate assessment in text-based tasks.

Just to be safe we also assess the reliability of GPT-4-turbo as a parser, we conducted a manual validation study:
\begin{itemize}
    \item We sampled 300 responses in total: 100 each from Last Letter, Shuffle Object, and GSM8K
    \item These responses were independently parsed by human evaluators.
    \item We then compared the human-parsed answers with those extracted by GPT-4-turbo.
\end{itemize}

The results of this validation are shown in Table \ref{tab:gpt4-validation}. These findings demonstrate an average alignment of 97.7\% between GPT-4-turbo and human-parsed answers, supporting our characterization of GPT-4-turbo as a near-perfect parser for this task.

\begin{table}[h]
\centering
\begin{tabular}{lc}
\hline
Task & GPT-4-Turbo correctness \\
\hline
Last Letter & 97/100 \\
Shuffle Obj & 96/100 \\
GSM8K & 100/100 \\
\hline
\end{tabular}
\caption{Alignment between GPT-4-turbo and human-parsed answers. In general we found GPT-4-turbo is very close to perfect parser which serves as a versatile parser to all kinds of task.}
\label{tab:gpt4-validation}
\end{table}

\begin{table}
\centering
\small
\begin{tabular}{lcc}
\hline
\textbf{Model} & \textbf{Regex Match} & \textbf{LLM Match} \\
\hline
GPT-3.5-Turbo & 43.7 & 75.5 \\
Gemini-1.5-Flash & 25.8 & 69.3 \\
Claude-3-Haiku & 67.4 & 85.8 \\
Gemma2-9b & 82.5 & 86.0 \\
LLaMA-3-8b & 46.9 & 55.7 \\
Mistral-7b-v0.3 & 10.4 & 52.4 \\
\hline
\end{tabular}
\caption{Comparison of model performance on regex match "\textit{The final answer is (\textbackslash d+)}" accuracy and using Claude-3-Haiku as answer parser.}
\label{tab:parser-comparison}
\end{table}

\section{Averaged numbers for all datasets}
\subsection{Zero shot prompting comparing Text, JSON, XML, YAML}
\label{app:zero_shot_results}

Table (\ref{tab:classification_tasks}, \ref{tab:reasoning_tasks}) shows all the number with standard deviation on all 4 format (NL, JSON, XML, YAML) in classification and reasoning tasks.

The JSON-mode scores for GPT 3.5 turbo, Gemini 1.5 Flash and LLaMA 3 8B are presented in Table \ref{tab:json_mode}. This table shows the performance of these three models on six different datasets when using JSON-mode.
\begin{table}
\centering
\caption{Zero shot prompting results for gemini-1.5-flash, gpt-3.5-turbo, claude-3-haiku, llama-3-8B, and gemma2-9B-IT averaged on 3 reasoning tasks with standard deviation in reasoning related task.}
\label{tab:reasoning_tasks}
\centering
\small
\begin{tabular}{@{}l@{\hspace{0.5em}}c@{\hspace{0.5em}}c@{\hspace{0.5em}}c@{}}
\toprule
& Last Letter & GSM8K & ShuffleObj \\
\midrule
Gemini-1.5-Flash & & & \\
\quad Text & 65.4 (3.1) & 89.3 (0.8) & 58.2 (13.0) \\
\quad JSON & 77.0 (7.3) & 89.2 (1.5) & 65.1 (5.3) \\
\quad XML & 74.2 (10.4) & 88.2 (2.2) & 50.4 (10.5) \\
\quad YAML & 71.4 (20.3) & 87.4 (3.7) & 34.3 (17.1) \\
\midrule
GPT-3.5 Turbo & & & \\
\quad Text & 56.7 (7.1) & 76.6 (2.8) & 20.4 (3.6) \\
\quad JSON & 25.2 (29.1) & 49.3 (12.0) & 20.9 (1.1) \\
\quad XML & 22.3 (27.8) & 45.1 (19.9) & 15.4 (1.8) \\
\quad YAML & 66.9 (22.0) & 73.9 (5.6) & 20.8 (1.3) \\
\midrule
Claude 3 Haiku & & & \\
\quad Text & 57.7 (21.1) & 86.5 (0.8) & 36.6 (8.2) \\
\quad JSON & 56.7 (16.7) & 23.4 (22.8) & 49.3 (4.8) \\
\quad XML & 33.8 (31.5) & 79.8 (7.0) & 48.1 (5.2) \\
\quad YAML & 31.6 (32.4) & 80.6 (2.8) & 18.1 (14.7) \\
\midrule
LLaMA 3 8B & & & \\
\quad Text & 70.1 (5.3) & 74.7 (0.6) & 27.0 (5.5) \\
\quad JSON & 28.0 (12.2) & 48.9 (6.7) & 15.7 (11.0) \\
\quad XML & 15.9 (4.8) & 56.7 (8.3) & 11.1 (3.6) \\
\quad YAML & 16.1 (10.4) & 46.1 (16.8) & 9.6 (3.6) \\
\midrule
Gemma2 9B IT & & & \\
\quad Text & 56.8 (9.8) & 86.5 (0.6) & 49.4 (5.8) \\
\quad JSON & 39.0 (6.8) & 84.2 (3.7) & 50.5 (8.9) \\
\quad XML & 43.7 (3.8) & 85.6 (0.6) & 38.5 (5.0) \\
\quad YAML & 23.4 (15.7) & 79.5 (4.1) & 23.0 (16.4) \\
\bottomrule
\end{tabular}
\end{table}

\begin{table*}[t!]
\centering
\caption{Zero shot prompting results for gemini-1.5-flash, gpt-3.5-turbo, claude-3-haiku, llama-3-8B, and gemma2-9B-IT averaged on 4 classification tasks with standard deviation in classification related task}
\label{tab:classification_tasks}
\centering
\small
\begin{tabular}{@{}l@{\hspace{0.5em}}c@{\hspace{0.5em}}c@{\hspace{0.5em}}c@{\hspace{0.5em}}c@{}}
\toprule
& DDXPlus & Sports & Task280 & MultiFin \\
\midrule
Gemini-1.5-Flash & & & & \\
\quad Text & 41.6 (6.6) & 79.9 (3.2) & 68.6 (2.5) & 63.5 (0.3) \\
\quad JSON & 60.3 (0.8) & 78.9 (1.3) & 70.3 (5.4) & 65.2 (1.1) \\
\quad XML & 59.4 (1.4) & 80.2 (0.7) & 70.0 (4.9) & 64.5 (1.6) \\
\quad YAML & 60.4 (1.6) & 80.1 (1.2) & 65.3 (12.7) & 64.1 (0.4) \\
\midrule
GPT-3.5 Turbo & & & & \\
\quad Text & 44.1 (3.2) & 67.2 (26.8) & 72.7 (6.3) & 63.0 (0.5) \\
\quad JSON & 55.5 (0.4) & 80.0 (3.3) & 70.6 (11.2) & 64.0 (0.9) \\
\quad XML & 53.0 (1.4) & 80.7 (1.1) & 66.2 (16.2) & 62.2 (1.1) \\
\quad YAML & 55.0 (0.8) & 80.9 (2.3) & 72.1 (8.0) & 65.4 (0.9) \\
\midrule
Claude 3 Haiku & & & & \\
\quad Text & 33.8 (13.5) & 77.8 (5.8) & 61.1 (11.0) & 62.0 (1.9) \\
\quad JSON & 52.0 (1.1) & 78.7 (2.8) & 49.5 (27.2) & 63.7 (1.3) \\
\quad XML & 50.8 (0.8) & 77.8 (3.8) & 45.0 (25.0) & 62.4 (1.1) \\
\quad YAML & 6.9 (5.3) & 76.4 (8.3) & 44.5 (24.2) & 61.8 (1.7) \\
\midrule
LLaMA 3 8B & & & & \\
\quad Text & 12.04 (15.2) & 69.49 (12.7) & 65.28 (3.4) & 60.26 (1.4) \\
\quad JSON & 23.37 (0.7) & 73.38 (3.5) & 39.46 (22.4) & 57.74 (2.0) \\
\quad XML & 11.35 (1.9) & 69.20 (5.5) & 35.36 (22.5) & 58.77 (3.2) \\
\quad YAML & 13.08 (4.1) & 68.25 (5.7) & 45.42 (24.4) & 49.74 (4.2) \\
\midrule
Gemma2 9B IT & & & & \\
\quad Text & 22.9 (5.8) & 76.1 (2.3) & 69.8 (7.7) & 70.0 (0.4) \\
\quad JSON & 53.0 (0.2) & 72.7 (1.6) & 65.6 (11.7) & 70.2 (0.7) \\
\quad XML & 52.9 (2.8) & 73.3 (2.4) & 68.1 (11.7) & 68.0 (0.7) \\
\quad YAML & 44.9 (2.2) & 73.0 (1.7) & 60.5 (11.0) & 69.8 (0.7) \\
\bottomrule
\end{tabular}
\end{table*}

\begin{table}[t]
\centering
\small
\begin{tabular}{@{}lcccc@{}}
\toprule
\textbf{Dataset} & \textbf{GPT3.5T} & \textbf{Gemini1.5F} & \textbf{LLaMA3 8B} \\
\midrule
LastLetter & 1.78 (0.3) & 0.67 (0.5) & \textbf{7.56 (2.7)} \\
GSM8K & 29.87 (0.8) & 47.78 (3.1) & \textbf{65.38 (1.3)} \\
\midrule
MultiFin & 66.00 (1.3) & \textbf{66.79 (0.4)} & 54.82 (1.5)\\
Sports & 76.82 (0.9) & \textbf{77.79 (0.4)} & 72.08 (2.6) \\
Task 280 & \textbf{78.07 (2.3)} & 67.19 (4.1) & 74.57 (2.0)\\
DDXPlus & 51.87 (2.8) & \textbf{84.92 (2.1)} & 22.59 (0.1)\\
\bottomrule
\end{tabular}
\caption{Averaged scores for JSON-mode to all 6 datasets, performance varies significantly across tasks and models, suggesting that different models may have strengths in different areas when using JSON-mode.}
\label{tab:json_mode}
\end{table}

\section{Prompt}
\label{app:prompt_format_details}
\subsection{Prompt Format}
For each task we fix the same template and only swapping the task description, format description, few shots example and question text.
\noindent
\begin{tcolorbox}[arc=10pt,outer arc=10pt]
    \textbf{Follow the instruction to complete the task:}\\
    \textcolor{blue}{\{task\_description\}}\\[1em]
    \textbf{Instruct:} \textcolor{blue}{\{format\_description\}}\\[1em]
    \textcolor{blue}{\{few shots\}}\\[1em]
    \textcolor{blue}{\{question\}}
  \label{fig:prompt_format}
\end{tcolorbox}
\noindent \textbf{Task Description} A task description describes the task and the final goal of the task.

\noindent \textbf{Format Description} A format description includes the target format (ie JSON, XML or YAML) and a targeted schema we intend the LLM response to adhere to.

\noindent For each description slot, we create 3 variations each which results in 9 prompt combinations. Each variation must retain the original meaning with slight change in wording, order of instruction. For each model we prompt all 9 prompts to calculate the sensitivity and variance of the final result.

If the current task requires reasoning, we include the zero shot chain-of-thought prompting : "Think step-by-step" in task description and ensures the LLM response to generate reasoning before giving the final answer.

\subsection{Prompt Variations}

Our study employs a range of prompt variations across multiple tasks to assess the robustness and generalizability of language models. We developed three distinct task description variations for each of the following datasets:

\begin{itemize}
    \item GSM8K (Figure \ref{fig:gsm8k_task_description})
    \item Last Letter (Figure \ref{fig:lastletter_task_description})
    \item Shuffle Object (Figure \ref{fig:shuffleobj_task_description})
    \item DDXPlus (Figure \ref{fig:ddxplus_task_description})
    \item Sports Understanding (Figure \ref{fig:sports_task_description})
    \item Natural Language - Task 280 (Figure \ref{fig:task280_task_description})
    \item MultiFin (Figure \ref{fig:multifin_task_description})
\end{itemize}

For tasks involving chain-of-thought reasoning (GSM8K, Last Letter, Shuffle Object Tracking, DDXPlus, Sports Understanding, and NL-Task 280), we implemented three prompt format variations. These are illustrated in Figures \ref{fig:gsm8k_prompt_description_1}, \ref{fig:gsm8k_prompt_description_2}, and \ref{fig:gsm8k_prompt_description_3}.

Additionally, we created three answering format variations for both reasoning-based tasks and those requiring direct answers. These "direct answer prompts" are presented in Figures \ref{fig:direct_answer_format1}, \ref{fig:direct_answer_format2}, and \ref{fig:direct_answer_format3}.

\begin{figure}[htbp]
\begin{minipage}{\columnwidth}
    \begin{mdframed}
    \textbf{Task description variation1:}\\
    You are a math tutor who helps students of all levels understand and solve mathematical problems.\\
  Read the last question carefully and think step by step before answering, the final answer must be only a number.\\
    \textbf{Task description variation2:}\\
      Read the last question carefully and think step by step before answering, the final answer must be only a number. You are a math tutor who helps students of all levels understand and solve mathematical problems.\\
    \textbf{Task description variation3:}\\
    Mathematical problem-solving task:\\
  • Given: A mathematical question or problem\\
  • Required: A numerical answer only\\
  • Role: You are a math tutor assisting students of all levels\\
  • Process: Think step by step to solve the problem\\
  Note: Read the question carefully before beginning your analysis.\\
    \end{mdframed}
\end{minipage}
\caption{GSM8K Task Description Variations}
\label{fig:gsm8k_task_description}
\end{figure}

\begin{figure}[htbp]
\begin{minipage}{\columnwidth}
    \begin{mdframed}
    \textbf{Task description variation1:}\\
    You are given a string of words and you need to take the last letter of each words and concate them.\\
  Read the last question carefully and think step by step before answering. \\
    \textbf{Task description variation2:}\\
      Read carefully for each of the last question and think step by step before answering. You are given a string of words and you need to take the last letter of each words and concatenate them.\\
    \textbf{Task description variation3:}\\
    String manipulation task:\\
  • Given: A sequence of words\\
  • Required: A new string made from the last letter of each word\\
  • Process: Think step by step to solve this challenge\\
  Note: Ensure you've read the question thoroughly before beginning.\\
    \end{mdframed}
\end{minipage}
\caption{Last Letter Task Description Variations}
\label{fig:lastletter_task_description}
\end{figure}

\begin{figure}[htbp]
\begin{minipage}{\columnwidth}
    \begin{mdframed}
    \textbf{Task description variation1:}\\
    In this task, you are tasked to answer the following commonsense knowledge task.\\
  Read carefully for each of the last question and think step by step before answering. \\
  Make sure the answer only contain one of these four choice : A, B, C, D, E, F, G\\
    \textbf{Task description variation2:}\\
  Read carefully for each of the last question and think step by step before answering. \\
  Make sure the answer only contain one of these four choice : A, B, C, D, E, F, G\\
  In this task, you are tasked to answer the following commonsense knowledge task.\\
    \textbf{Task description variation3:}\\
  Context understanding assessment:\\
  • Given: A story related to many person in the same place\\
  • Required: Determine if the person who is in the end of the story\\
  • Process: Think step by step to analyze the context\\
  • Output: Answer the correct answer and only contain one of these seven choice : A, B, C, D, E, F, G\\
\end{mdframed}
\end{minipage}
\caption{Shuffle object Task Description Variations}
\label{fig:shuffleobj_task_description}
\end{figure}

\begin{figure}[htbp]
\begin{minipage}{\columnwidth}
    \begin{mdframed}
    \textbf{Task description variation1:}\\
    Extract the following RESPONSE final answer, your answer should be the one which match any of these valid diagnoses:\\
    - Possible NSTEMI / STEMI\\
    - Spontaneous rib fracture\\
    - Pulmonary embolism\\
    - Pulmonary neoplasm\\
    ... \\
    - Scombroid food poisoning\\
    RESPONSE:\\
    \textbf{Task description variation2:}\\
    Act as a medical doctor and diagnose the patient based on the given patient profile\\
    All possible valid diagnoses for you to choose from are as follows:\\
    - Possible NSTEMI / STEMI\\
    - Spontaneous rib fracture\\
    - Pulmonary embolism\\
    - Pulmonary neoplasm\\
    ... \\
    - Scombroid food poisoning\\
    \textbf{Task description variation3:}\\
  Medical diagnosis task:\\
  • Given: A patient profile\\
  • Required: Diagnose the patient based on the provided information\\
  • Process: Think step by step to analyze the patient's symptoms and history\\
  • Output: Select one diagnosis from the provided list of valid options\\
  Note: Carefully review the patient profile and the list of possible diagnoses before making your determination. Do not answer "Insufficient information" - you must choose from the given options.\\
  Valid diagnoses (select one):\\
  - Possible NSTEMI / STEMI\\
  - Spontaneous rib fracture\\
  - Pulmonary embolism\\
  - Pulmonary neoplasm\\
  ...\\
  - Scombroid food poisoning\\
\end{mdframed}
\end{minipage}
\caption{DDXPlus Task Description Variations}
\label{fig:ddxplus_task_description}
\end{figure}

\begin{figure}[htbp]
\begin{minipage}{\columnwidth}
    \begin{mdframed}
    \textbf{Task description variation1:}\\
  You are given a sentence and your task is to determine whether a sentence relating to sports is plausible or implausible\\
  Read carefully for each of the last question and think step by step before answering. \\
  Answer yes if its plausible, no if implausible\\
    \textbf{Task description variation2:}\\
You are given a sentence and your task is to determine whether a sentence relating to sports is plausible or implausible. Read carefully for each of the last question and think step by step before answering. Answer yes if its plausible, no if implausible\\
    \textbf{Task description variation3:}\\
  Sentence plausibility assessment:\\
  • Given: A sentence related to sports\\
  • Required: Determine if the sentence is plausible or implausible\\
  • Process: Think step by step to analyze the sentence\\
  • Output: Answer "yes" if plausible, "no" if implausible\\
  \end{mdframed}
\end{minipage}
\caption{Sports Task Description Variations}
\label{fig:sports_task_description}
\end{figure}

\begin{figure}[htbp]
\begin{minipage}{\columnwidth}
    \begin{mdframed}
    \textbf{Task description variation1:}\\
In this task, you are given a short passage that conveys stereotype or anti-stereotype about a specific target. A stereotype is an over-generalized belief about a particular group of people. An anti-stereotype is an idea that goes against a common stereotype. You are expected to classify the passage into four types of stereotype or anti-stereotype: gender, profession, race, and religion.\\
    \textbf{Task description variation2:}\\
  You are expected to classify the passage into four types of stereotype or anti-stereotype: gender, profession, race, and religion.\\
  In this task, you are given a short passage that conveys stereotype or anti-stereotype about a specific target. A stereotype is an over-generalized belief about a particular group of people. An anti-stereotype is an idea that goes against a common stereotype.\\
    \textbf{Task description variation3:}\\
  Sentence stereotype assessment:\\
  • Given: A passage related to stereotype or anti-stereotype\\
  • Required: Determine if the paragraph is one of these four category : gender, profession, race, and religion\\
  • Output: Answer only one of the four category\\
  \end{mdframed}
\end{minipage}
\caption{Task 280 Task Description Variations}
\label{fig:task280_task_description}
\end{figure}

\begin{figure}[htbp]
\begin{minipage}{\columnwidth}
    \begin{mdframed}
    \textbf{Task description variation1:}\\
    Act as a finance expert and assign the content based to the valid category\\
    All possible valid category for you to choose from are as follows (one category per line, in the format of <category>):\\
    - Finance\\
    - Technology\\
    - Tax and Accounting\\
    - Business and Management\\
    - Government and Controls\\
    - Industry\\
    Your answer MUST based on the above options, do not answer Insufficient information\\    \textbf{Task description variation2:}\\
  Act as a finance expert and assign the content based to the valid category\\
    Your answer MUST based on the above options, do not answer Insufficient information\\
    All possible valid category for you to choose from are as follows (one category per line, in the format of <category>):\\
    - Finance\\
    - Technology\\
    - Tax and Accounting\\
    - Business and Management\\
    - Government and Controls\\
    - Industry\\
    \textbf{Task description variation3:}\\
 Act as a finance expert and assign the content based to the valid category\\
    All possible valid category for you to choose from are as follows (one category per line, in the format of <category>):\\
    * Finance\\
    * Technology\\
    * Tax and Accounting\\
    * Business and Management\\
    * Government and Controls\\
    * Industry\\
    Your answer MUST based on the above options, do not answer Insufficient information\\
  \end{mdframed}
\end{minipage}
\caption{MultiFin Task Description Variations}
\label{fig:multifin_task_description}
\end{figure}

\begin{figure}[htbp]
\begin{minipage}{\columnwidth}
    \begin{mdframed}
    \textbf{DA prompt description variation 1:}\\
    \textbf{Natural language:}\\
Derive the most likely category to answer key.
    Provide your output in the following valid text format:\\
    Answer: ...\\
    \textbf{JSON:}\\
    Derive the most likely category to answer key.
    Provide your output in the following valid JSON format:\\
    ```json\\
    \{\\
      "answer": "..."\\
    \}\
    ```\\
    \textbf{YAML:}\\
    Derive the most likely category to answer key.
    Provide your output in the following valid YAML format:\\
    ```yaml\\
    answer: ...\\
    ```\\
    \textbf{XML:}\\
    Derive the most likely category to answer block
    Provide your output in the following valid YAML format:\\
    ```xml\\
    <root>\\
      <answer>...</answer>\\
    </root>\\
    ```\\
    \end{mdframed}
\end{minipage}
\caption{Variation 1 for direct Answering format with only answer field in all 4 format.}
\label{fig:direct_answer_format1}
\end{figure}

\begin{figure}[htbp]
\begin{minipage}{\columnwidth}
    \begin{mdframed}
    \textbf{DA prompt description variation 2:}\\
    \textbf{Natural language:}\\
    Provide your output in the following text format:\\
    Step by step reasoning: ... \\
    Answer: The final answer is ...\\
    \textbf{JSON:}\\
    Provide your output in the following valid JSON format:\\
    ```json\\
    \{\\
        "step\_by\_step\_reasoning": ...,\\
        "answer": ... \\
    \}\\
    ```\\
    \textbf{YAML:}\\
    Provide your output in the following valid YAML format:\\
    ```yaml\\
    step\_by\_step\_reasoning: |\\
      ...\\
    answer: ...\\
    ```\\
    \textbf{XML:}\\
    Provide your output in the following valid XML format:\\
    ```xml\\
    <root>\\
      <step\_by\_step\_reasoning>...\\
      </step\_by\_step\_reasoning>\\
      <answer>...</answer>\\
    </root>\\
    ```\\
    \end{mdframed}
\end{minipage}
\caption{Variation 2 for direct Answering format with only answer field in all 4 format.}
\label{fig:direct_answer_format2}
\end{figure}

\begin{figure}[htbp]
\begin{minipage}{\columnwidth}
    \begin{mdframed}
    \textbf{DA prompt description variation 3:}\\
    \textbf{Natural language:}\\
    Provide your output in the following text format:\\
    Answer: <think step by step>. The final answer is <answer>\\
    \textbf{JSON:}\\
    Provide your output in the following valid JSON format:\\
    ```json\\
    \{\\
        "reason": "<think step by step>",\\
        "answer": <answer>\\
    \}\\
    ```\\
    \textbf{YAML:}\\
    Provide your output in the following valid YAML format:\\
    ```yaml\\
    reasoning: |\\
      <think step by step>,\\
    answer: <answer>\\
    ```\\
    \textbf{XML:}\\
    Provide your output in the following valid XML format:\\
    ```xml\\
    <root>\\
      <reason>[think step by step]</reason>\\
      <answer>[answer]</answer>\\
    </root>\\
    ```\\
    \end{mdframed}
\end{minipage}
\caption{Variation 3 for direct Answering format with only answer field in all 4 format.}
\label{fig:direct_answer_format3}
\end{figure}

\begin{figure}[htbp]
\begin{minipage}{\columnwidth}
    \begin{mdframed}
    \textbf{CoT prompt description variation 1:}\\
    \textbf{Natural language:}\\
    Provide your output in the following text format:\\
    Answer: <reasoning first>. The final answer is <answer>\\
    \textbf{JSON:}\\
    Provide your output in the following valid JSON format:\\
    ```json\\
    \{\\
        "reason": ...,\\
        "answer": ...\\
    \}\\
    ```\\
    \textbf{YAML:}\\
    Provide your output in the following valid YAML format:\\
    ```yaml\\
    reasoning: |\\
      ...\\
    answer: ...\\
    ```\\
    \textbf{XML:}\\
    Provide your output in the following valid XML format:\\
    ```xml\\
    <root>\\
      <reason>...</reason>\\
      <answer>...</answer>\\
    </root>\\
    ```\\
    \end{mdframed}
\end{minipage}
\caption{Reasoning response prompt - Variation 1}
\label{fig:gsm8k_prompt_description_1}
\end{figure}

\begin{figure}[htbp]
\begin{minipage}{\columnwidth}
    \begin{mdframed}
    \textbf{CoT prompt description variation 2:}\\
    \textbf{Natural language:}\\
    Provide your output in the following text format:\\
    Step by step reasoning: ... \\
    Answer: The final answer is ...\\
    \textbf{JSON:}\\
    Provide your output in the following valid JSON format:\\
    ```json\\
    \{\\
        "step\_by\_step\_reasoning": ...,\\
        "answer": ...\\
    \}\\
    ```\\
    \textbf{YAML:}\\
    Provide your output in the following valid YAML format:\\
    ```yaml\\
    step\_by\_step\_reasoning: |\\
      ...\\
    answer: ...\\
    ```\\
    \textbf{XML:}\\
    Provide your output in the following valid XML format:\\
    ```xml\\
    <root>\\
      <step\_by\_step\_reasoning>...\\
      </step\_by\_step\_reasoning>\\
      <answer>...</answer>\\
    </root>\\
    ```\\
    \end{mdframed}
  
\end{minipage}
\caption{Reasoning response prompt - Variation 2}
\label{fig:gsm8k_prompt_description_2}
\end{figure}

\begin{figure}[htbp]
\begin{minipage}{\columnwidth}
    \begin{mdframed}
    \textbf{CoT prompt description variation 3:}\\
    \textbf{Natural language:}\\
    Provide your output in the following text format:\\
    Answer: <think step by step>. The final answer is <answer>\\
    \textbf{JSON:}\\
    Provide your output in the following valid JSON format:\\
    ```json\\
    \{\\
        "reason": "<think step by step>",\\
        "answer": <answer>\\
    \}\\
    ```\\
    \textbf{YAML:}\\
    Provide your output in the following valid YAML format:\\
    ```yaml\\
    reasoning: |\\
      <think step by step>,\\
    answer: <answer>\\
    ```\\
    \textbf{XML:}\\
    Provide your output in the following valid XML format:\\
    ```xml\\
    <root>\\
      <reason>[think step by step]</reason>\\
      <answer>[answer]</answer>\\
    </root>\\
    ```\\
    \end{mdframed}
\end{minipage}
\caption{Reasoning response prompt - Variation 3}
\label{fig:gsm8k_prompt_description_3}
\end{figure}

\end{document}